\documentclass[]{fairmeta}
\usepackage{makecell}
\usepackage{wrapfig}
\usepackage{tabularx}
\usepackage{textcomp}
\usepackage{stfloats}
\usepackage{url}
\usepackage{verbatim}
\usepackage{titlesec}
\usepackage{tocloft}
\usepackage{adjustbox}
\usepackage{multirow}
\usepackage{pifont}
\usepackage[sc]{mathpazo}
\usepackage{tikz}
\usepackage{comment}
\usepackage{amsmath,amssymb}
\usepackage{colortbl}
\usepackage{natbib}
\usepackage{color}
\usepackage{booktabs} 
\usepackage{hyperref}
\usepackage{graphicx}
\usepackage{subcaption}
\usepackage{multirow}
\usepackage{subcaption}
\RequirePackage{xspace}
\makeatletter
\DeclareRobustCommand\onedot{\futurelet\@let@token\@onedot}
\def\@onedot{\ifx\@let@token.\else.\null\fi\xspace}
\usepackage[most]{tcolorbox}
\usepackage{xcolor}
\usepackage{array}
\usepackage{tabularx}
\usepackage{siunitx} 
\usepackage{makecell}
\usepackage[table]{xcolor}
\usepackage{caption}
\definecolor{headerpurple}{HTML}{d8d2fc}
\definecolor{rowgray}{gray}{0.95}

\makeatother

\definecolor{adptorange}{RGB}{248, 205, 172}
\definecolor{cmpblue}{RGB}{189, 215, 238}
\definecolor{cmpblue}{RGB}{189, 215, 238}

\definecolor{our_red}{RGB}{232,157,160}
\definecolor{our_blue}{RGB}{136,206,230}
\definecolor{our_orange}{RGB}{246,200,168}
\definecolor{our_green}{RGB}{178,211,164}

\definecolor{attn_code0}{RGB}{247,215,200}
\definecolor{attn_code1}{RGB}{238,169,139}
\definecolor{mlp_code0}{RGB}{204,201,221}
\definecolor{mlp_code1}{RGB}{102,95,153}
\definecolor{mygray}{HTML}{f0f0f0}

\definecolor{token_blue}{RGB}{84, 120, 140}

\usepackage{pifont}
\usepackage{bbding}
\usepackage{fontawesome}
\usepackage{xspace}

\usepackage{float}

\newlength\savewidth

\newcolumntype{x}[1]{>{\centering\arraybackslash}p{#1pt}}
\newcolumntype{y}[1]{>{\raggedright\arraybackslash}p{#1pt}}
\newcolumntype{z}[1]{>{\raggedleft\arraybackslash}p{#1pt}}

\renewcommand{\paragraph}[1]{\vspace{1mm}\noindent\textbf{#1}}
\usepackage{colortbl}
\usepackage{xcolor}
\usepackage{wrapfig}

\renewcommand{\paragraph}[1]{\vspace{1.25mm}\noindent\textbf{#1}}

\usepackage{algorithm}
\usepackage{listings}

\definecolor{codeblue}{rgb}{0.25, 0.5, 0.5}
\definecolor{codekw}{rgb}{0.35, 0.35, 0.75}
\lstdefinestyle{Pytorch}{
    language = Python,
    backgroundcolor = \color{white},
    basicstyle = \fontsize{9pt}{8pt}\selectfont\ttfamily\bfseries,
    columns = fullflexible,
    aboveskip=1pt,
    belowskip=1pt,
    breaklines = true,
    captionpos = b,
    commentstyle = \color{codeblue},
    keywordstyle = \color{codekw},
}

\definecolor{green}{HTML}{009000}
\definecolor{red}{HTML}{ea4335}

\title{WildWorld: A Large-Scale Dataset for Dynamic World Modeling with Actions and Explicit State toward Generative ARPG}
\author[1,2,4,*,\S]{Zhen Li}
\author[1,3,*,\S]{Zian Meng}
\author[1]{Shuwei Shi}
\author[5]{Wenshuo Peng}
\author[2,4,\dagger]{Yuwei Wu}
\author[1]{Bo Zheng}
\author[1,\dagger]{Chuanhao Li}
\author[1,\dagger]{Kaipeng Zhang}

\affiliation[1]{Alaya Studio, Shanda AI Research Tokyo}
\affiliation[2]{Beijing Institute of Technology}
\affiliation[3]{Shanghai Innovation Institute\\}
\affiliation[4]{Shenzhen MSU-BIT University}
\affiliation[5]{Tsinghua University}

\abstract{
Dynamical systems theory and reinforcement learning view world evolution as latent-state dynamics driven by actions, with visual observations providing partial information about the state.
Recent video world models attempt to learn this action-conditioned dynamics from data.
However, existing datasets rarely match the requirement:
they typically lack diverse and semantically meaningful action spaces,
and actions are directly tied to visual observations rather than mediated by underlying states.
As a result, actions are often entangled with pixel-level changes, making it difficult for models to learn structured world dynamics and maintain consistent evolution over long horizons.
In this paper, we propose WildWorld, a large-scale action-conditioned world modeling dataset with explicit state annotations, automatically collected from a photorealistic AAA action role-playing game (\textit{Monster Hunter: Wilds}).
WildWorld contains over 108 million frames and features more than 450 actions, including movement, attacks, and skill casting, together with synchronized per-frame annotations of character skeletons, world states, camera poses, and depth maps.
We further derive WildBench to evaluate models through Action Following and State Alignment.
Extensive experiments reveal persistent challenges in modeling semantically rich actions and maintaining long-horizon state consistency, highlighting the need for state-aware video generation.
\textcolor{red}{We are looking for researchers, engineers and interns interested in world models and AI-native games.}
}
\github{\url{https://shandaai.github.io/wildworld-project/}}
\Code{\url{https://github.com/ShandaAI/WildWorld}}
\date{\today} 

\begin{document}
\maketitle

\renewcommand{\thefootnote}{\fnsymbol{footnote}}
\footnotetext[4]{This work was done during the internship at Shanda AI Research Tokyo.}
\footnotetext[1]{Equal contribution.}
\footnotetext[2]{Corresponding authors: wuyuwei@bit.edu.cn; chuanhao.li@shanda.com; kaipeng.zhang@shanda.com}

% \teaser{
\begin{figure}[!h]
    \centering
  \includegraphics[width=1.0\linewidth]{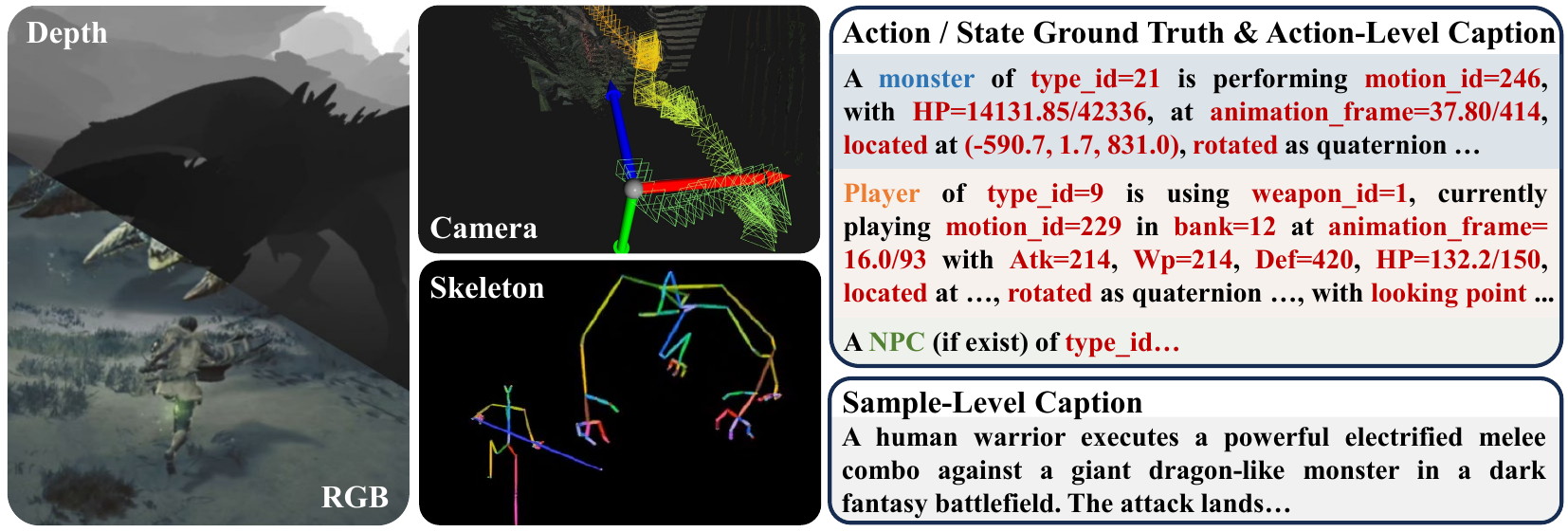}\vspace{5pt}
  \captionof{figure}{We present a large-scale dataset curated from game engines for dynamic world modeling. It contains RGB frames with aligned depth maps, camera poses, skeleton, and action / state ground truth. We provide both fine-grained action-level captions and sample-level captions, making the dataset applicable to various experimental settings.}
  \label{fig:teaser}
\end{figure}
% }
% % teasor

\section{Introduction}
Understanding and predicting how the world evolves from observations is one of the central goals of artificial intelligence \cite{schmidhuber2015learning,ha2018world,kim2020active}.
Both dynamical systems theory \cite{bertsekas2012dynamic,hafner2023mastering} and reinforcement learning \cite{sutton2018reinforcement} typically model the world as a latent-state dynamical process, where the environment evolves through state transitions driven by actions. From this perspective, visual observations are merely partial and noisy projections of the true system state.
Therefore, learning a predictive model of the world requires inferring latent states and modeling their action-conditioned state transitions.
Such world models are crucial for enabling agents to plan, reason, and interact with complex environments over long horizons.

Recent years have witnessed significant progress in video generation and world models \cite{wan2025wan,hacohen2026ltx,team2026advancing}.
Many recent approaches~\cite{matrixgame2,ji2025memflow,yume1_5} attempt to learn environment dynamics from large-scale video datasets by training generative models that predict future frames conditioned on past observations and actions.
However, despite the increasing capability of such models, existing datasets remain insufficient for effectively learning structured action-conditioned dynamics.
Most existing datasets provide only simple action annotations with limited semantic meaning, such as basic movements or camera rotations \cite{sekai,spatialvid}.
Moreover, the effects of these actions are often directly observable in the visual observations.
For example, the action “move left” is typically reflected in the video as a corresponding change in viewpoint.

However, in many cases, actions are not defined through explicit observation variations but instead manifest through implicit state transitions.
For instance, the action ``shoot'' implicitly affects internal state variables such as the ``remaining ammunition count''.
This state cannot be reliably inferred from visual observations alone, yet it plays a crucial role in determining future visual outcomes.
When the remaining ammunition reaches zero, executing the shoot action will no longer produce firing effects or projectiles, leading to visual results that differ significantly from those observed when ammunition is available.
Such coupling makes it difficult for models to disentangle state transitions from observation variations, thereby hindering the learning of stable and interpretable world dynamics.
As a result, current models often perform poorly in long-horizon prediction tasks, where small errors accumulate over time and eventually lead to noticeable inconsistencies or instability in the generated results.

In this paper, we propose WildWorld, a large-scale video dataset for action-conditioned world modeling with explicit state annotations.
The dataset is automatically collected from the photorealistic AAA action role-playing game \textit{Monster Hunter: Wilds}. WildWorld features a rich and semantically meaningful action space containing over 450 actions, including movement, attacks, and skill casting.
To facilitate data collection, we develop a bespoke toolchain capable of recording per-frame ground-truth annotations, including player actions, character skeletons, world states, camera poses, depth maps, etc.
The toolchain is integrated with an automated gameplay pipeline, allowing the dataset to scale easily to over 108M frames of gameplay footage while covering diverse interactive scenarios.
By capturing complex interactions and the underlying state transitions, WildWorld enables the study of long-horizon compositional action sequences and their effects on evolving world states, providing a valuable foundation for building, training, and systematically evaluating state-aware interactive world models.

Furthermore, we derive WildBench, a benchmark constructed from WildWorld for evaluating interactive world models. WildBench introduces two key evaluation metrics: Action Following and State Alignment.
Specifically, Action Following measures the agreement between generated videos and the ground-truth sub-actions.%, which is automatically evaluated using large vision-language models.
State alignment quantitatively measures the accuracy of state transitions by tracking skeletal keypoints in the generated videos and comparing them with the corresponding ground-truth annotations.
We design several baseline models for state-aware video generation and compare them with existing approaches on WildBench.
The experimental results reveal the limitations of current models and provide insights for future research, particularly in improving state transition modeling and long-horizon consistency.

To summarize, our contributions are threefold:
(1) We propose \textbf{WildWorld}, a large-scale video dataset comprising over 108M frames, with a rich action space and diverse frame-level ground truth annotations, including player actions, character skeletons, world states, camera poses, depth, etc.
(2) We curate \textbf{WildBench}, a benchmark for evaluating interactive world models, featuring two carefully designed metrics: Action Following and State Alignment.
(3) We conduct extensive experiments and analysis on WildBench, which provide insights into the future development of interactive world models.

\section{Related Work}
\subsection{Interactive World Models}

Recent advances in video generation models \cite{sora2,longlive,hacohen2026ltx} have enabled the development of interactive world generation models \cite{genie,cosmos,wan2025wan}.
In the realm of video generation,
text-to-video \cite{chen2024videocrafter2,li2024t2v,yume} and image-to-video generation \cite{xing2024dynamicrafter,xu2024easyanimate,shi2024motion,wan2025wan} have achieved remarkable progress in generation quality and temporal consistency.
As for interactive video generation,
works \cite{yume1_5,genie3,team2026advancing} enable interaction by switching prompts during the generation process,
while works \cite{genie3,hyworld2025,gao2025longvie, matrixgame1,matrixgame2,hunyuangame,yan,voyager} introduce actions via keyboard control \cite{matrixgame2,genie3} and camera poses \cite{camctrl1, camctrl2} on top of image-to-video generation to control the generated video.
Despite producing promising results, these methods are limited by a restricted action space and tightly couple action control with pixel-level video changes.
In contrast, we focus on state-aware video generation via action control, which features a rich action space (over 450 actions) and aims to use states as an intermediate representation to convey the effects of actions on pixel-level video generation.

Some recent works \cite{wang2026mechanistic, yue2025simulating, garrido2026learning, lillemarkflow, 3d_as_code} attempt to introduce latent state representations into video generation models to better capture environment dynamics.
However, these approaches typically represent the world state as an implicit latent variable learned from visual observations.
By contrast, we focus on explicit, semantically meaningful states and introduce WildWorld, a large-scale dataset with state annotations for learning and analyzing state dynamics.

\subsection{Video Generation Dataset}

Recent progress in video generation has been driven by several large-scale datasets, such as OpenVid-1M \cite{nan2024openvid}, MiraData \cite{ju2024miradata}, Open-Sora \cite{lin2024open}, and SpatialVID \cite{spatialvid}, which provide large collections of internet videos for training generative models. 
More recent works have begun to explore datasets for world modeling or interactive video generation, including OmniWorld \cite{zhou2025omniworld}, Sekai \cite{sekai}, GF-Minecraft \cite{yu2025gamefactory}, PLAICraft \cite{he2025plaicraft}, and GameGen-X \cite{che2024gamegen}.
While these datasets introduce gameplay videos or action signals to capture environment dynamics, they still primarily rely on visual observations and lack explicit, semantically meaningful state representations.
MIND \cite{ye2026mind} proposes a benchmark for evaluating memory consistency and action control in world models.
Compared to the above works, WildWorld provides explicit state annotations, including character skeletons, world states, camera poses, and depth, enabling models to learn structured state dynamics and supporting direct evaluation of state alignment and action following.
\section{WildWorld Dataset}

\begin{figure*}[t]
\centering
\includegraphics[width=0.9\linewidth]{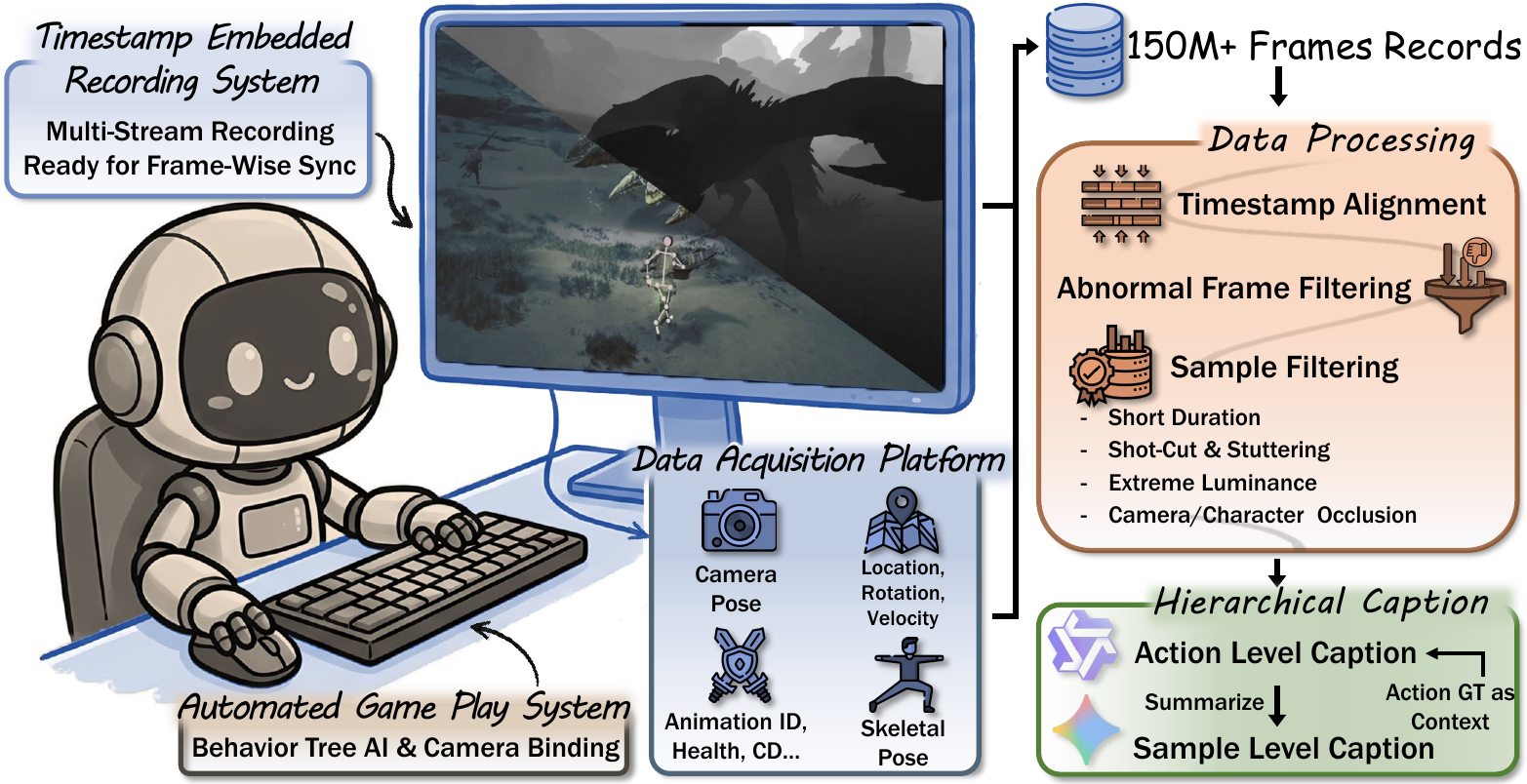}
\caption{The WildWorld dataset curation pipeline.}
\label{fig:framework}
\end{figure*}

The overall process of curating the WildWorld dataset includes four major parts: data acquisition platform, automated gameplay pipeline, data processing and caption annotation pipeline, see \cref{fig:framework} for illustrations.

\subsection{Data Acquisition Platform}

In the data acquisition stage, we collect the interaction data required for training and evaluating interactive world models, organized into three categories: actions, states, and observations.         
Actions specify the control inputs that drive interactions, states describe the underlying evolution of the game world, and observations correspond to its visual manifestations.
These three types of data can be recorded at different stages of modern game execution.
In \textit{Monster Hunter: Wilds}, the game engine processes player inputs, maintains and updates the world state. 
While the rendering pipeline consumes information from the game engine to produce the final imagery.
Following this separation, we develop a dedicated game data acquisition platform engineered for the high fidelity record of various categories of data.

Specifically, we build our data acquisition platform to record both player actions and world states ground truth,
including the executed actions, absolute location, rotation, and velocities of the player character and monsters in the game world, their current animation IDs, and gameplay attributes such as health, stamina/mana-like resources as actions and states.
We additionally record the skeletal poses of both the player character and monsters.
For world observations, we instrument the rendering pipeline to record RGB frames, depth maps, and the intrinsic and extrinsic parameters of the in-game camera. We further remove the HUD by disabling the corresponding late stage shaders.
This yields clean, HUD free frames that better reflect the game world for training and evaluating interactive world models.

\subsection{Automated Game Record Pipeline}

Turning the captured raw streams into a usable and scalable dataset requires solving several challenges at the system level. 
On one hand, to enable long running collection with minimal human intervention, we implement automated gameplay system, including menu UI navigation and player action execution.
On the other hand, to recorded different interaction data captured by separate tools, we design a robust recording system with embedded timestamps, to facilitate subsequent cross source synchronization and alignment.

\noindent
\textbf{Automated Game Play System}.
\textit{Monster Hunter: Wilds} follows a quest-based structure in which each session tasks a party of up to four characters, one player-controlled protagonist and three NPC companions, with hunting one or two large monsters.
Our automation consists of two components.
For quest selection, we invoke the game engine's UI components to programmatically navigate in-game menus and randomly sample quest-NPC combinations, ensuring diverse coverage over maps, monsters, and team compositions.
For automated combat, we leverage the game's built-in rule-based companion AI.
We enable automated combat by leveraging the behavior trees that drive NPC companions to fight autonomously, and correspondingly adjust the in-game camera binding so that the entire party can act without human input.
A natural concern is whether rule-based AI yields overly repetitive behavior.
We argue that the resulting trajectories remain sufficiently diverse for two reasons.
First, the combinatorial action space is large: the AI must select among dozens of moves and continuously adjust timing and positioning in response to monster behavior, which itself is stochastic.
Second, the interaction between multiple AI-controlled characters and a reactive monster creates a high-dimensional dynamical system whose trajectories vary substantially across sessions, even under the same scripted logic.
During automated combat, the camera is managed by the game's native target-lock system, which dynamically adjusts the camera position and angle to keep the engaged monster within the field of view while maintaining visual stability.

\noindent
\textbf{Recording System}. We develop a recording system for the simultaneous recording of interaction data from multiple sources. 
For structured information represented in text form, such as actions and states, recording is straightforward.
At each engine tick, these interaction data are uniformly recorded, serialized in JSON format, and written to one local file.
Given that the full screen is typically occupied by the RGB frame in standard rendering setups, image frame such as RGB and depth require a different strategy for simultaneous record.
To achieve that, we develop a dedicated system based on OBS Studio and Reshade.
Specifically, a custom Reshade shader partitions the full display into four sub-windows, two of which present the RGB and depth frames from the rendering buffer.
In practice, we set the full display resolution to 2K, yielding sub-windows of 720p.
We further adapt a modified version of OBS Studio to simultaneously record different sub-windows of the screen as separate recording streams, allowing RGB and depth to be recorded with different encoding settings.
Specifically, RGB is recorded with lossy HEVC compression under variable bitrate control, using a target bitrate of 16 Mbps and a maximum of 20 Mbps, so as to reduce storage cost while maintaining high visual quality.
In contrast, depth is recorded losslessly to preserve geometric precision and avoid discontinuities caused by lossy compression.
In practice, we use the HEVC encoder with B frames enabled, and the resulting bitrate of the depth stream remains around 20 Mbps.
In addition, we embed timestamp information into the recordings of multiple sources, which serves as a unified basis to process them and obtain synchronized data samples in the next section. 

\subsection{Data Processing and Annotation Pipeline}

While the temporally tagged multi source recordings collected in the previous stage contain rich action, state, and observation data, they may still contain misalignment, duplicated or dropped frames caused by occasional runtime instability, and low quality or uninformative content such as occlusions and cutscenes, making them unsuitable for direct use by interactive world models.
We therefore apply a set of multi-dimensional filters to remove low-quality samples.
Based on the resulting samples, we further annotate hierarchical caption to support fine-grained modeling and evaluation.

\noindent
\textbf{Sample Filtering}.
We filter the samples along the following dimensions to improve the overall data quality.
\begin{itemize}
    \item \textbf{Duration Filtering}.
    Very short samples provide limited value for interactive world modeling. 
    We therefore discard samples shorter than 81 frames.
    
    \item \textbf{Temporal Continuity Filtering}.
    Given the state record in every frame of samples is associated with a timestamp, we can directly measure the temporal gap between adjacent frames. 
    Excessive gaps typically indicate either stuttering in the game or recording system, or transitions into non-combat content such as cutscenes. 
    The latter can be identified in our data, since our platform only records data during combat or travel.
    We discard any sample in which the gap between two adjacent frames exceeds 1.5 times the target frame interval, \textit{i.e.}, approximately 50\,ms at 30 FPS.

    \item \textbf{Luminance Filtering}.
    Overly bright or dark visuals in games can create visually distinctive gameplay experiences, for example in combat effects or in nighttime scenes.
    However, such samples are less suitable for stable model training.
    We apply a simple filter based on the luma channel in YUV color space of the RGB frames, and remove samples with more than 15 consecutive frames of extremely high or low average brightness.

    \item \textbf{Camera Occlusion Filtering}.
    We remove samples with foreground occlusion, such as rocks, trees, or other scene geometry blocking the character. 
    We detect such cases using the spring-arm behavior of the third-person camera: when occlusion occurs, the arm contracts, leading to an abnormally small camera–character distance; 
    therefore, we discard samples whose recorded distances fall below a threshold for a sustained number of frames.
    We further exclude samples with abrupt player position changes, such as fast travel, as they break visual continuity.

    \item \textbf{Character Occlusion Filtering}.
    Severe character overlap in the first frame cam introduce ambiguity into image to video generation.
    We identify inter character overlap by projecting 3D skeletal keypoints onto screen coordinates in the first frame and discarding samples in which the overlap area between characters exceeds 30\% of either character's projected area.
\end{itemize}

\noindent
\textbf{Hierarchical Caption Annotations}
Fine-grained captions are important for capturing interaction details and enabling the training of more precisely controllable models, for example through prompt switching~\cite{ji2025memflow,yang2025longlive}.
Leveraging the action annotations provided in WildWorld, we segment each sample into action sequences according to the frame wise action IDs, such that the action remains unchanged within each sequence, \textit{e.g.}, walking forward or charging a heavy attack.
For each sequence, we sample RGB frames at 1 FPS, resize them to 480p, and use Qwen3-VL-235B-A22B-Instruct served with vLLM to generate detailed captions.
To compensate for the model’s limited familiarity with game specific scenarios, we additionally include the corresponding action and state ground-truth in the prompt context.
We further provide sample level captions by summarizing all action sequence captions in one sample with Gemini 3 Flash.

\subsection{Dataset Statistics}

After processing and filtering, we yield WildWorld with 108 million frames and 119 annotation columns per frame.

\begin{figure*}[t]
\centering
\includegraphics[width=\linewidth]{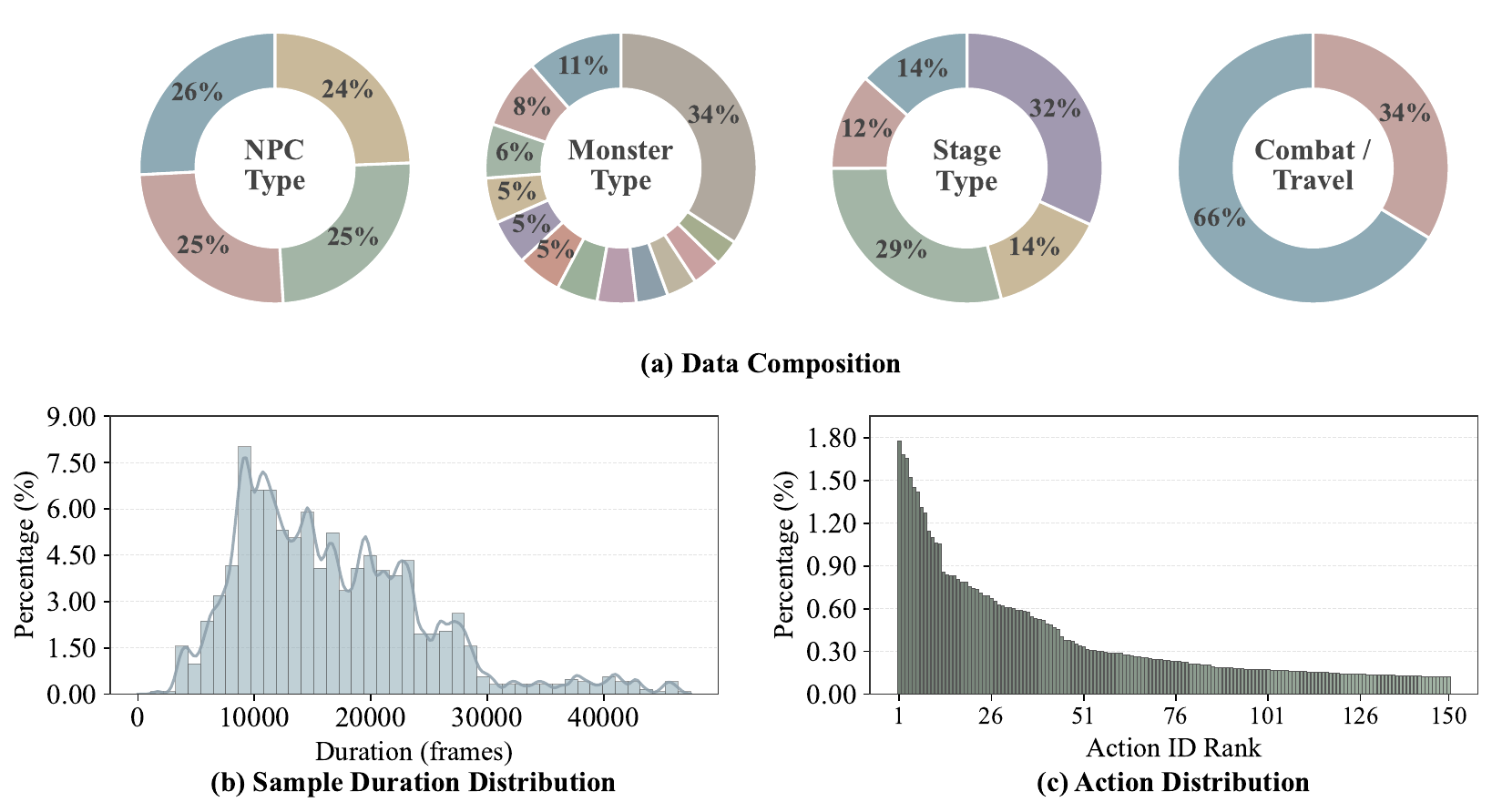}
\caption{Wildworld dataset statistics overview. (a)~Data composition by character type, monster species, stage, and combat / travel ratio. (b)~Distribution of sample durations in frames. (c)~Frequency distribution of the top-150 action IDs, exhibiting a long-tail pattern.}
\label{fig:diversity_stats}
\end{figure*}

\noindent
\textbf{Entity Diversity}.
The dataset covers 29 unique monster species, 4 player characters, and 4 weapon types (Great Sword, Long Sword, Bow, Dual Blades).
As shown in \cref{fig:diversity_stats}\,(a), character types and weapon types are near-uniformly distributed, while monster species follow a long-tailed distribution dominated by a few frequent targets.
Multi-monster encounters also appear, with 7 secondary species present in the data.
This diversity is significant for training world models that generalize across entities and interaction patterns.

\noindent
\textbf{Scene Complexity}.
Gameplay spans 5 distinct stages set in an open-world map with diverse environments including deserts, snowy mountains, forests, swamps, and wastelands, under varying weather (sunny, rainy) and time-of-day (day, night) conditions.
As shown in \cref{fig:diversity_stats}\,(a), approximately 66\% of clips capture active combat, while the remaining 34\% depict traversal on mounts, providing a broad range of interaction contexts for training and evaluating world models.

\noindent
\textbf{Temporal and Spatial Dynamics}.
\cref{fig:diversity_stats}\,(b) shows the distribution of sample durations.
The majority of clips span 4,000 to 28,000 frames, while a smaller subset exceeds 40,000 frames (over 30 minutes of gameplay), capturing extended combat sequences or exploration that demand long-horizon consistency.
Spatially, the camera-to-character distance has a median of 15.69 units and the character-to-monster distance a median of 12.63 units. These close proximities ensure that the character and monster are prominently featured in the video frames, with clear visibility of their actions and state changes.

\noindent
\textbf{Action Richness}.
Each frame's character state is encoded as a (weapon type, bank ID, motion ID) triplet, yielding 5,960 unique character action triplets across 24 banks and 455 motion IDs.
These actions span movement, attacks, evasion, defense, item usage, and inter-action transitions, covering the full range of in-game interactions.
Monsters exhibit 2,132 unique action pairs across 13 banks and 527 motion IDs.
\cref{fig:diversity_stats}\,(c) shows the frequency of the top-150 character action IDs, which account for 58.49\% of all samples and follow a long-tail distribution, indicating rich behavioral variety.

\section{WildBench Benchmark}

Evaluating interactive world models requires measuring not only visual plausibility, but also how well the model follows input actions and produces aligned states.
Leveraging the action and state ground-truth provided by WildWorld, we evaluate generated videos from four perspectives: video quality, camera control, action following, and state alignment.
Among these, action following and state alignment more directly evaluate how well a model follows input actions and responds with aligned states under interaction, both of which are central to interactive world modeling.
This distinguishes our benchmark from existing ones~\cite{li2025worldmodelbench,duan2025worldscore,ye2026mind}, which mainly emphasize perceptual quality, controllability, or physics plausibility.

\subsection{Evaluation Metric}

We comprehensively evaluate interactive world models from four perspectives: 

\noindent
\textbf{Video Quality} characterizes the overall perceptual quality of generated videos in terms of both motion and appearance. 
We evaluate this dimension using four VBench metrics~\cite{huang2024vbench}: Motion Smoothness (MS) assesses the smoothness and physical plausibility of generated motion; Dynamic Degree (DD) measures the magnitude of motion to penalize overly static videos; Aesthetic Quality (AQ) reflects the perceived artistic and visual appeal of the generated content; Image Quality (IQ) evaluates low-level visual distortions, such as over-exposure, noise, and blur.

\noindent
\textbf{Camera Control} is essential for interactive world models, as inaccurate viewpoint control can prevent the intended observations from being properly presented.
We quantitatively evaluate camera control by measuring the discrepancy between the ground-truth camera trajectories and the camera trajectories estimated from generated videos using a structure from motion model following CameraCtrl~\cite{he2024cameractrl}.
To reduce the impact of scale mismatch between trajectories from the game engine and those estimated from video, we apply a scalar alignment factor to the estimated translations before evaluation.
We then compute Absolute Trajectory Error (ATE) and Relative Pose Error (RPE) for both translation and rotation~\cite{sturm2012benchmark}.
ATE measures the absolute deviation from the ground-truth trajectory, reflecting the overall accuracy of camera control, while RPE measures the discrepancy between relative motions and is therefore more sensitive to local consistency and accumulated drift along the trajectory.
In practice, we estimate camera trajectories using ViPE~\cite{huang2025vipe}.

\noindent
\textbf{Action Following} evaluates whether the model responds to input actions with the corresponding behaviors in generated videos.
Since each sample may contain multiple actions, we perform evaluation at the action sequence level for finer-grained assessment.
Based on the frame-wise action ID annotations in WildWorld, we divide each sample into action segments within which the action remains unchanged.
For each segment, we extract the corresponding frame range from both the generated video and the ground-truth video, and use Gemini 3 Flash to judge whether they express the same action.
We further group actions into three categories, namely movement, fast displacement, and attack, based on their action IDs, and design detailed prompts for each category.
Each segment is assigned a score of 1 if the generated and ground-truth clips are judged to be consistent, and 0 otherwise.
The final score is the average over all segments.

\noindent
\textbf{State Alignment}.
We use the poses of the player character and monsters as a proxy for state, since pose directly reflects many underlying world states and can also indirectly reveal others, such as the death pose when health reaches zero.
Using the ground-truth skeletons in WildWorld, we extract key skeletal points and project them onto screen coordinates to obtain 2D trajectories for each sample.
For generated videos, we focus on settings based on image-to-video generation, where the first frame is the ground-truth.
Thus, we initialize keypoints from the first frame and track them in generated videos with TAPNext~\cite{zholus2025tapnext}. 
Then, we define the State Alignment score as the mean coordinate accuracy between the predicted and ground-truth trajectories over all keypoints.
For each keypoint, the coordinate accuracy is computed as the average fraction of frames whose predicted locations fall within thresholds of 4, 8, 16, and 32 pixels from the ground truth.
We note that while state evolution may be stochastic due to factors such as random events, alignment with the ground truth remains statistically meaningful over a number of samples.

\subsection{Data Curation}

It is worth noting that all samples in WildWorld dataset are, in principle, available for evaluation, allowing users to flexibly construct custom test sets based on scenario, difficulty, and other factors.
In this paper, we manually curate a representative set of 200 samples covering diverse difficulty levels, combat scenarios, character and monster types, and events such as skill usage, knockdowns, deaths, and critical hits.
Among them, 100 samples involve cooperation between the player and NPCs against monsters, while the other 100 consist of one on one combat between the player and a monster.

\section{Experiments and Analysis}

In this section, we first assess the reliability of the proposed benchmark metrics and their alignment with human preference.
We then train different interactive world modeling approaches on WildWorld and evaluate them on WildBench using the ground truth annotations provided in WildWorld dataset.

\subsection{Compared Approaches}

\noindent
\textbf{Camera-Conditioned Video Generation}.
In this setting, model takes a camera trajectory, an initial image, and a text prompt as inputs to generate a video that follows the defined camera motion.
We fine-tune the Wan2.2-Fun-5B-Control-Camera~\cite{wan2025wan} model with camera trajectories ground-truth in WildWorld dataset, the resulting model is dubbed as \textit{\textbf{CamCtrl}}.
The baseline model uses a rule-based approach to convert discrete camera control action inputs into
camera poses, from which it computes Plücker embeddings~\cite{plkemb} for each frame and injects them into
the model.
In contrast, we directly use the ground-truth per-frame camera poses in WildWorld as inputs for fine-tuning.

\noindent
\textbf{Skeleton-Conditioned Video Generation}.
Skeletal pose provides a direct and fine-grained representation of character motion.
We introduce a skeleton-conditioned setting that takes the first frame and a skeleton video as inputs.
We fine-tune Wan2.2-Fun-5B-Control~\cite{wan2025wan}, a video-to-video model that supports skeleton-based pose videos as a control signal, and dub the resulting model \textbf{\textit{SkelCtrl}}.
To construct the skeleton video, we use the per-frame 3D skeleton keypoints and their joint-tree structure annotated in the WildWorld dataset, project them into screen coordinate under the ground-truth camera pose, and render them as a colored-skeleton video matching the input format expected by the baseline model. \par\nopagebreak[4]

\noindent
\textbf{State-Conditioned Video Generation}.
Based on \textit{\textbf{CamCtrl}}, we design a state-aware model: \textit{\textbf{StateCtrl}}, which injects states into the video generation process.
We first perform structured modeling of different states, dividing them into discrete states (\textit{e.g.}, monster type, weapon category) and continuous states (\textit{e.g.}, coordinates, health).
Discrete states are mapped to vector representations through trainable embedding, while continuous states are encoded into the same feature space using an MLP.
At the encoding stage, we adopt a hierarchical modeling strategy with entity-level and global-level representations.
Each entity (\textit{e.g.}, monster) encodes its own states, while global states (such as recording time) are also incorporated.
We use transformer \cite{vaswani2017attention} architecture to model the relationships between entities, producing a unified state embedding representation.
The resulting embedding is aligned with video frames and injected into the intermediate layers of DiT as a conditioning signal to guide the video generation process.
In addition, we introduce a state decoder to recover the state information from the state embedding representation, a state predictor to predict the next-frame state.
During training, a decoder loss is used to ensure that the embedding preserves the original states;
a predictor loss is used to supervise the state predictor, enhancing the temporal consistency and predictability of the state representation.
During inference, \textit{\textbf{StateCtrl}} supports using only the ground-truth state of the first frame, while the states of subsequent frames are autoregressively predicted by the state predictor.
We denote this model as \textbf{\textit{StateCtrl-AR}}.

\noindent
Across all settings, models are trained at a resolution of $544 \times 960$ with 81 frames per sample and a frame rate of 16 FPS, using a batch size of 1 and a learning rate of $1\times10^{-5}$.
Training is performed for 250,000 iterations with a batch size of 8 using the Adam optimizer.
During inference, we adopt the same resolution and frame rate, and use 50 sampling steps.

\begin{table*}[t]
\centering
\renewcommand{\arraystretch}{1.3}
\setlength{\tabcolsep}{5.0pt}
\caption{Comparison of different interactive video generation approaches trained on WildWorld and evaluated on WildBench. Lower is better for ATE and RPE; higher is better for the others.}
\begin{tabular}{l|cccc|cc|c|c}
\hline
\multirow{2}{*}{Method} 
& \multicolumn{4}{c|}{Video Quality} 
& \multicolumn{2}{c|}{Camera Control} 
& \multicolumn{1}{c|}{Action} 
& \multicolumn{1}{c}{State} \\
\cline{2-5}\cline{6-7}
& MS 
& DD 
& AQ 
& IQ
& ATE($\downarrow$)
& RPE($\downarrow$)
& \multicolumn{1}{c|}{Following}
& \multicolumn{1}{c}{Alignment} \\
\hline
Baseline & 96.38 & 99.00 & 50.81 & 65.62 & 4.63 & 0.18 & 53.77 & 11.29 \\
CamCtrl & 97.85 & 97.00 & 48.29 & 62.88 & 2.02 & 0.13 & 83.46 & 15.18 \\
SkelCtrl & 97.85 & 95.00 & 47.92 & 62.43 & 2.55 & 0.10 & 92.81 & 22.03 \\
StateCtrl & 97.45 & 99.00 & 50.86 & 67.78 & 0.94 & 0.07 & 85.66 & 16.06 \\
StateCtrl-AR & 97.43 & 99.00 & 50.90 & 67.76 & 1.01 & 0.08 & 74.66 & 16.13 \\
\hline
\end{tabular}
\label{tab:main_results}
\end{table*}

\subsection{Overall Evaluation}

\noindent
\textbf{Evaluation of the Proposed Benchmark Metrics}.
We validate the reliability of the proposed Action Following and State Alignment metrics, as well as their alignment with human preference.
For Action Following, we use the same evaluation protocol with human judgments instead of model-based assessment. 
Specifically, 10 volunteers are recruited, and each segment is annotated by three volunteers.
Segments with inconsistent annotations are discarded, accounting for about 5\% of the full set. 
We then measure the consistency between human judgments and model scores.
For State Alignment, we run keypoint tracking directly on ground truth videos and evaluate the resulting trajectories using the same protocol to verify the reliability of the metric.

The experimental results show that human judgments and the proposed model-based Action Following metric achieve 85\% agreement on WildBench, indicating that the metric can reliably reflect human evaluation of action consistency.
Moreover, the State Alignment metric achieves 43.23\% coordinate accuracy against the  skeleton ground-truth, demonstrating its effectiveness in measuring alignment between generated and ground-truth state evolution.

\noindent
\textbf{Evaluation of the Interactive Video Generation}. Table~\ref{tab:main_results} presents the WildBench evaluation results of different interactive world model approaches trained on the WildWorld dataset. Here, Baseline refers to Wan2.2-TI2V-5B~\cite{wan2025wan}, and Video Quality is reported as a percentage. We observe that

\begin{itemize}
    \item \textbf{All approaches improve over the baseline on interaction-related metrics}.
    For example, the camera-conditioned video generation model \textit{\textbf{CamCtrl}} reduces ATE and RPE on Camera Control by $\triangle$2.61 and $\triangle$0.05, respectively. 
    Taking skeletal video as conditional input, \textit{\textbf{SkelCtrl}} achieves nearly 100\% average improvement on Action Following and State Alignment.
    Moreover, directly conditioning on discrete and continuous state information, \textit{\textbf{StateCtrl}} and \textit{\textbf{StateCtrl-AR}} improve performance across all three metrics.
    This highlights the usefulness of our WildWorld dataset, as diverse approaches can benefit from it.

    \item \textbf{VBench appear saturated on WildWorld.} We observe that all methods achieve over 95\% on the MS (Motion Smoothness) and DD (Dynamic Degree) metrics under Video Quality.
    In contrast, their actual ability to produce reasonable motion and dynamics still differs substantially, as evidenced by the large performance difference on Action Following and State Alignment.
    This suggests that interactive world models require more fine-grained and nuanced evaluation metrics to assess highly dynamic video generation, which aligns with the design goal of WildBench.
    
    \item \textbf{Directly using visual signals as conditional input yield a trade-off.}
    We find that \textit{\textbf{SkelCtrl}}, which uses visual signals for interactive control input, achieves larger gains on interaction-related metrics than \textit{\textbf{StateCtrl}}, which learns soft embeddings from the same information; however, it comes at the cost of lower video quality, as reflected by decreased AQ (Aesthetic Quality) and IQ (Image Quality) scores under the Video Quality evaluation. We further analyze this pattern in the qualitative evaluation blow.

    \item \textbf{Autoregressive interactive world models show promise}.
    Using only the first-frame state and autoregressively predicting subsequent states as control inputs, \textit{\textbf{StateCtrl-AR}} achieves performance comparable to \textit{\textbf{StateCtrl}}, but exhibits a noticeable drop in \textbf{Action Following}.
    We attribute this degradation to error accumulation in iterative next-state prediction, a phenomenon also observed in autoregressive video generation~\cite{huang2025self,longlive,ji2025memflow}.
    We believe this paradigm can be combined with autoregressive video generation and may even further advance its development.

\end{itemize}

\begin{figure*}[t]
\centering
\includegraphics[width=\linewidth]{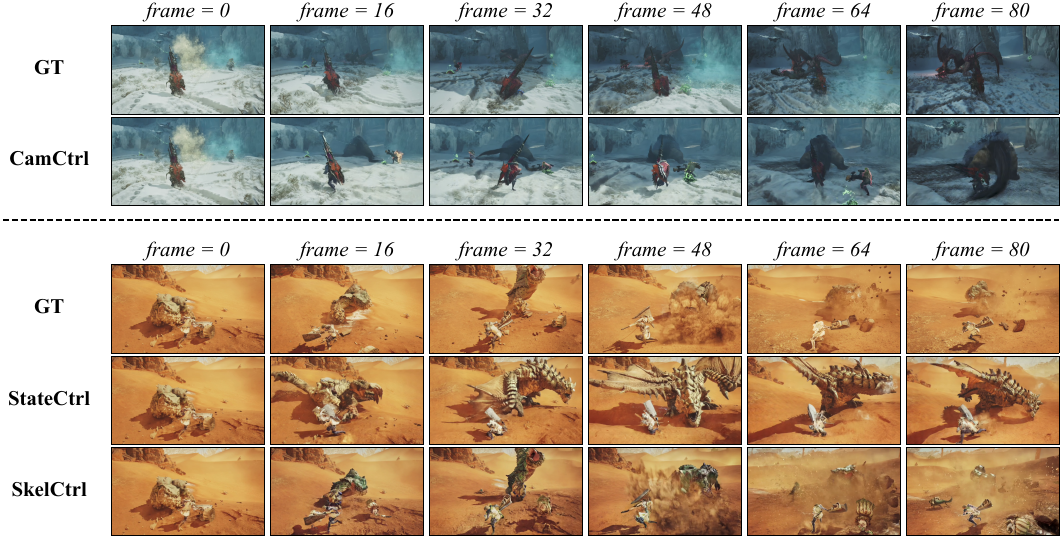}
\caption{Qualitative comparisons of different interactive world modeling approaches trained on the WildWorld dataset.}
\label{fig:visualization}
\end{figure*}

\subsection{Qualitative Evaluation}

Table~\ref{fig:visualization} presents visual comparisons of different interactive world modeling approaches on two test examples.
The top example shows that \textit{\textbf{CamCtrl}} produces camera motion consistent with the ground truth, but fails to capture the monster’s dynamics.
In particular, for the bottom example, we observe that \textit{\textbf{StateCtrl}} generates a clearer foreground subject, whereas in the ground truth the subject is partially occluded by splashing sand and gravel; in contrast, \textit{\textbf{SkelCtrl}} better reproduces this effect. This observation is consistent with the abovementioned stronger image-quality performance of \textit{\textbf{StateCtrl}}, as clearer frames are typically perceived as having higher image quality.

\section{Conclusion}
In this paper, We have presented WildWorld, a large-scale video dataset with explicit state annotations to facilitate the study of action-conditioned world modeling.
The dataset is automatically collected from a photorealistic AAA action role-playing game, \textit{Monster Hunter: Wilds}, through a scalable data collection pipeline.
WildWorld provides a rich and meaningful action space with over 450 actions, and each video is annotated with frame-level annotations including character skeletons, world states, camera poses, and depth.
Furthermore, we have introduced WildBench, a benchmark derived from WildWorld, which enables the quantitative evaluation of action following and state alignment.
Experimental results demonstrate that existing models still face significant challenges in modeling semantically rich actions and maintaining long-horizon state consistency.
These findings highlight the importance of incorporating explicit state information for advancing action-conditioned video generation and world modeling.

\bibliographystyle{abbrv}
\bibliography{references}

@article{ji2025memflow,
  title={Memflow: Flowing adaptive memory for consistent and efficient long video narratives},
  author={Ji, Sihui and Chen, Xi and Yang, Shuai and Tao, Xin and Wan, Pengfei and Zhao, Hengshuang},
  journal={arXiv preprint arXiv:2512.14699},
  year={2025}
}

@article{yang2025longlive,
  title={Longlive: Real-time interactive long video generation},
  author={Yang, Shuai and Huang, Wei and Chu, Ruihang and Xiao, Yicheng and Zhao, Yuyang and Wang, Xianbang and Li, Muyang and Xie, Enze and Chen, Yingcong and Lu, Yao and others},
  journal={arXiv preprint arXiv:2509.22622},
  year={2025}
}

@inproceedings{duan2025worldscore,
  title={Worldscore: A unified evaluation benchmark for world generation},
  author={Duan, Haoyi and Yu, Hong-Xing and Chen, Sirui and Fei-Fei, Li and Wu, Jiajun},
  booktitle={Proceedings of the IEEE/CVF International Conference on Computer Vision},
  pages={27713--27724},
  year={2025}
}

@article{li2025worldmodelbench,
  title={Worldmodelbench: Judging video generation models as world models},
  author={Li, Dacheng and Fang, Yunhao and Chen, Yukang and Yang, Shuo and Cao, Shiyi and Wong, Justin and Luo, Michael and Wang, Xiaolong and Yin, Hongxu and Gonzalez, Joseph E and others},
  journal={arXiv preprint arXiv:2502.20694},
  year={2025}
}

@article{ye2026mind,
  title={MIND: Benchmarking Memory Consistency and Action Control in World Models},
  author={Ye, Yixuan and Lu, Xuanyu and Jiang, Yuxin and Gu, Yuchao and Zhao, Rui and Liang, Qiwei and Pan, Jiachun and Zhang, Fengda and Wu, Weijia and Wang, Alex Jinpeng},
  journal={arXiv preprint arXiv:2602.08025},
  year={2026}
}

@inproceedings{huang2024vbench,
  title={Vbench: Comprehensive benchmark suite for video generative models},
  author={Huang, Ziqi and He, Yinan and Yu, Jiashuo and Zhang, Fan and Si, Chenyang and Jiang, Yuming and Zhang, Yuanhan and Wu, Tianxing and Jin, Qingyang and Chanpaisit, Nattapol and others},
  booktitle={Proceedings of the IEEE/CVF Conference on Computer Vision and Pattern Recognition},
  pages={21807--21818},
  year={2024}
}

@article{he2024cameractrl,
  title={Cameractrl: Enabling camera control for text-to-video generation},
  author={He, Hao and Xu, Yinghao and Guo, Yuwei and Wetzstein, Gordon and Dai, Bo and Li, Hongsheng and Yang, Ceyuan},
  journal={arXiv preprint arXiv:2404.02101},
  year={2024}
}

@article{huang2025vipe,
  title={Vipe: Video pose engine for 3d geometric perception},
  author={Huang, Jiahui and Zhou, Qunjie and Rabeti, Hesam and Korovko, Aleksandr and Ling, Huan and Ren, Xuanchi and Shen, Tianchang and Gao, Jun and Slepichev, Dmitry and Lin, Chen-Hsuan and others},
  journal={arXiv preprint arXiv:2508.10934},
  year={2025}
}

@inproceedings{zholus2025tapnext,
  title={Tapnext: Tracking any point (tap) as next token prediction},
  author={Zholus, Artem and Doersch, Carl and Yang, Yi and Koppula, Skanda and Patraucean, Viorica and He, Xu Owen and Rocco, Ignacio and Sajjadi, Mehdi SM and Chandar, Sarath and Goroshin, Ross},
  booktitle={Proceedings of the IEEE/CVF International Conference on Computer Vision},
  pages={9693--9703},
  year={2025}
}

@inproceedings{sturm2012benchmark,
  title={A benchmark for the evaluation of RGB-D SLAM systems},
  author={Sturm, J{\"u}rgen and Engelhard, Nikolas and Endres, Felix and Burgard, Wolfram and Cremers, Daniel},
  booktitle={2012 IEEE/RSJ international conference on intelligent robots and systems},
  pages={573--580},
  year={2012},
  organization={IEEE}
}

@book{sutton2018reinforcement,
  title={Reinforcement Learning: An Introduction},
  author={Sutton, Richard S. and Barto, Andrew G.},
  year={2018},
  publisher={MIT Press}
}

@article{hafner2023mastering,
  title={Mastering diverse domains through world models},
  author={Hafner, Danijar and Pasukonis, Jurgis and Ba, Jimmy and Lillicrap, Timothy},
  journal={arXiv preprint arXiv:2301.04104},
  year={2023}
}

@book{bertsekas2012dynamic,
  title={Dynamic programming and optimal control: Volume I},
  author={Bertsekas, Dimitri},
  volume={4},
  year={2012},
  publisher={Athena scientific}
}

@article{ha2018world,
  title={World models},
  author={Ha, David and Schmidhuber, J{\"u}rgen},
  journal={arXiv preprint arXiv:1803.10122},
  volume={2},
  number={3},
  pages={440},
  year={2018}
}

@article{schmidhuber2015learning,
  title={On learning to think: Algorithmic information theory for novel combinations of reinforcement learning controllers and recurrent neural world models},
  author={Schmidhuber, J{\"u}rgen},
  journal={arXiv preprint arXiv:1511.09249},
  year={2015}
}

@inproceedings{kim2020active,
  title={Active world model learning with progress curiosity},
  author={Kim, Kuno and Sano, Megumi and De Freitas, Julian and Haber, Nick and Yamins, Daniel},
  booktitle=ICML,
  pages={5306--5315},
  year={2020},
  organization={PMLR}
}

@article{wan2025wan,
  title={Wan: Open and advanced large-scale video generative models},
  author={Wan, Team and Wang, Ang and Ai, Baole and Wen, Bin and Mao, Chaojie and Xie, Chen-Wei and Chen, Di and Yu, Feiwu and Zhao, Haiming and Yang, Jianxiao and others},
  journal={arXiv preprint arXiv:2503.20314},
  year={2025}
}

@article{gao2025longvie,
  title={LongVie 2: Multimodal Controllable Ultra-Long Video World Model},
  author={Gao, Jianxiong and Chen, Zhaoxi and Liu, Xian and Zhuang, Junhao and Xu, Chengming and Feng, Jianfeng and Qiao, Yu and Fu, Yanwei and Si, Chenyang and Liu, Ziwei},
  journal={arXiv preprint arXiv:2512.13604},
  year={2025}
}

@article{hyworld2025,
  title={HY-World 1.5: A Systematic Framework for Interactive World Modeling with Real-Time Latency and Geometric Consistency},
  author={Team HunyuanWorld},
  journal={arXiv preprint},
  year={2025}
}

@article{team2026advancing,
  title={Advancing Open-source World Models},
  author={Team, Robbyant and Gao, Zelin and Wang, Qiuyu and Zeng, Yanhong and Zhu, Jiapeng and Cheng, Ka Leong and Li, Yixuan and Wang, Hanlin and Xu, Yinghao and Ma, Shuailei and others},
  journal={arXiv preprint arXiv:2601.20540},
  year={2026}
}

@article{sekai,
  title={Sekai: A Video Dataset towards World Exploration},
  author={Li, Zhen and Li, Chuanhao and Mao, Xiaofeng and Lin, Shaoheng and Li, Ming and Zhao, Shitian and Xu, Zhaopan and Li, Xinyue and Feng, Yukang and Sun, Jianwen and others},
  journal={arXiv preprint arXiv:2506.15675},
  year={2025}
}

@article{spatialvid,
  title={Spatialvid: A large-scale video dataset with spatial annotations},
  author={Wang, Jiahao and Yuan, Yufeng and Zheng, Rujie and Lin, Youtian and Gao, Jian and Chen, Lin-Zhuo and Bao, Yajie and Zhang, Yi and Zeng, Chang and Zhou, Yanxi and others},
  journal={arXiv preprint arXiv:2509.09676},
  year={2025}
}

@article{plkemb,
  title={Light field networks: Neural scene representations with single-evaluation rendering},
  author={Sitzmann, Vincent and Rezchikov, Semon and Freeman, Bill and Tenenbaum, Josh and Durand, Fredo},
  journal={Advances in Neural Information Processing Systems},
  volume={34},
  pages={19313--19325},
  year={2021}
}

@misc{sora2,
    title={Sora 2},
    author={Sora 2 Contributors},
    howpublished = {\url{https://openai.com/index/sora-2/}},
    year={2025}
}

@inproceedings{yume1_5,
  title={Yume-1.5: A Text-Controlled Interactive World Generation Model},
  author={Mao, Xiaofeng and Li, Zhen and Li, Chuanhao and Xu, Xiaojie and Ying, Kaining and He, Tong and Pang, Jiangmiao and Qiao, Yu and Zhang, Kaipeng},
  journal=CVPR,
  year={2026}
}

@article{yume,
  title={Yume: An interactive world generation model},
  author={Mao, Xiaofeng and Lin, Shaoheng and Li, Zhen and Li, Chuanhao and Peng, Wenshuo and He, Tong and Pang, Jiangmiao and Chi, Mingmin and Qiao, Yu and Zhang, Kaipeng},
  journal={arXiv preprint arXiv:2507.17744},
  year={2025}
}

@article{vaswani2017attention,
  title={Attention is all you need},
  author={Vaswani, Ashish and Shazeer, Noam and Parmar, Niki and Uszkoreit, Jakob and Jones, Llion and Gomez, Aidan N and Kaiser, {\L}ukasz and Polosukhin, Illia},
  journal={Advances in neural information processing systems},
  volume={30},
  year={2017}
}

@misc{genie3,
    title={Genie 3},
    author={Genie 3 Contributors},
    howpublished = {\url{https://deepmind.google/models/genie/}},
    year={2025}
}

@inproceedings{camctrl1,
  title={Cameractrl: Enabling camera control for video diffusion models},
  author={He, Hao and Xu, Yinghao and Guo, Yuwei and Wetzstein, Gordon and Dai, Bo and Li, Hongsheng and Yang, Ceyuan},
  booktitle=ICLR,
  year={2025}
}

@InProceedings{camctrl2,
    author    = {He, Hao and Yang, Ceyuan and Lin, Shanchuan and Xu, Yinghao and Wei, Meng and Gui, Liangke and Zhao, Qi and Wetzstein, Gordon and Jiang, Lu and Li, Hongsheng},
    title     = {CameraCtrl II: Dynamic Scene Exploration via Camera-controlled Video Diffusion Models},
    booktitle = ICCV,
    month     = {October},
    year      = {2025},
    pages     = {13416-13426}
}

@article{hunyuangame,
  title={Hunyuan-GameCraft: High-dynamic Interactive Game Video Generation with Hybrid History Condition},
  author={Li, Jiaqi and Tang, Junshu and Xu, Zhiyong and Wu, Longhuang and Zhou, Yuan and Shao, Shuai and Yu, Tianbao and Cao, Zhiguo and Lu, Qinglin},
  journal={arXiv preprint arXiv:2506.17201},
  year={2025}
}

@article{longlive,
  title={Longlive: Real-time interactive long video generation},
  author={Yang, Shuai and Huang, Wei and Chu, Ruihang and Xiao, Yicheng and Zhao, Yuyang and Wang, Xianbang and Li, Muyang and Xie, Enze and Chen, Yingcong and Lu, Yao and others},
  journal={arXiv preprint arXiv:2509.22622},
  year={2025}
}

@article{yan,
  title={Yan: Foundational interactive video generation},
  author={Ye, Deheng and Zhou, Fangyun and Lv, Jiacheng and Ma, Jianqi and Zhang, Jun and Lv, Junyan and Li, Junyou and Deng, Minwen and Yang, Mingyu and Fu, Qiang and others},
  journal={arXiv preprint arXiv:2508.08601},
  year={2025}
}

@article{matrixgame1,
  title={Matrix-Game: Interactive World Foundation Model},
  author={Zhang, Yifan and Peng, Chunli and Wang, Boyang and Wang, Puyi and Zhu, Qingcheng and Kang, Fei and Jiang, Biao and Gao, Zedong and Li, Eric and Liu, Yang and others},
  journal={arXiv preprint arXiv:2506.18701},
  year={2025}
}

@article{matrixgame2,
  title={Matrix-game 2.0: An open-source, real-time, and streaming interactive world model},
  author={He, Xianglong and Peng, Chunli and Liu, Zexiang and Wang, Boyang and Zhang, Yifan and Cui, Qi and Kang, Fei and Jiang, Biao and An, Mengyin and Ren, Yangyang and others},
  journal={arXiv preprint arXiv:2508.13009},
  year={2025}
}

@article{voyager,
  title={Voyager: Long-Range and World-Consistent Video Diffusion for Explorable 3D Scene Generation},
  author={Huang, Tianyu and Zheng, Wangguandong and Wang, Tengfei and Liu, Yuhao and Wang, Zhenwei and Wu, Junta and Jiang, Jie and Li, Hui and Lau, Rynson WH and Zuo, Wangmeng and others},
  journal={arXiv preprint arXiv:2506.04225},
  year={2025}
}

@inproceedings{chen2024videocrafter2,
  title={Videocrafter2: Overcoming data limitations for high-quality video diffusion models},
  author={Chen, Haoxin and Zhang, Yong and Cun, Xiaodong and Xia, Menghan and Wang, Xintao and Weng, Chao and Shan, Ying},
  booktitle={Proceedings of the IEEE/CVF conference on computer vision and pattern recognition},
  pages={7310--7320},
  year={2024}
}

@article{li2024t2v,
  title={T2v-turbo: Breaking the quality bottleneck of video consistency model with mixed reward feedback},
  author={Li, Jiachen and Feng, Weixi and Fu, Tsu-Jui and Wang, Xinyi and Basu, Sugato and Chen, Wenhu and Wang, William Yang},
  journal={Advances in neural information processing systems},
  volume={37},
  pages={75692--75726},
  year={2024}
}

@inproceedings{xing2024dynamicrafter,
  title={Dynamicrafter: Animating open-domain images with video diffusion priors},
  author={Xing, Jinbo and Xia, Menghan and Zhang, Yong and Chen, Haoxin and Yu, Wangbo and Liu, Hanyuan and Liu, Gongye and Wang, Xintao and Shan, Ying and Wong, Tien-Tsin},
  booktitle={European Conference on Computer Vision},
  pages={399--417},
  year={2024},
  organization={Springer}
}

@article{xu2024easyanimate,
  title={Easyanimate: A high-performance long video generation method based on transformer architecture},
  author={Xu, Jiaqi and Zou, Xinyi and Huang, Kunzhe and Chen, Yunkuo and Liu, Bo and Cheng, MengLi and Shi, Xing and Huang, Jun},
  journal={arXiv preprint arXiv:2405.18991},
  year={2024}
}

@inproceedings{shi2024motion,
  title={Motion-i2v: Consistent and controllable image-to-video generation with explicit motion modeling},
  author={Shi, Xiaoyu and Huang, Zhaoyang and Wang, Fu-Yun and Bian, Weikang and Li, Dasong and Zhang, Yi and Zhang, Manyuan and Cheung, Ka Chun and See, Simon and Qin, Hongwei and others},
  booktitle={ACM SIGGRAPH 2024 Conference Papers},
  pages={1--11},
  year={2024}
}

@inproceedings{genie,
  title={Genie: Generative interactive environments},
  author={Bruce, Jake and Dennis, Michael D and Edwards, Ashley and Parker-Holder, Jack and Shi, Yuge and Hughes, Edward and Lai, Matthew and Mavalankar, Aditi and Steigerwald, Richie and Apps, Chris and others},
  booktitle={Forty-first International Conference on Machine Learning},
  year={2024}
}

@article{cosmos,
  title={Cosmos world foundation model platform for physical ai},
  author={Agarwal, Niket and Ali, Arslan and Bala, Maciej and Balaji, Yogesh and Barker, Erik and Cai, Tiffany and Chattopadhyay, Prithvijit and Chen, Yongxin and Cui, Yin and Ding, Yifan and others},
  journal={arXiv preprint arXiv:2501.03575},
  year={2025}
}

@article{hacohen2026ltx,
  title={LTX-2: Efficient Joint Audio-Visual Foundation Model},
  author={HaCohen, Yoav and Brazowski, Benny and Chiprut, Nisan and Bitterman, Yaki and Kvochko, Andrew and Berkowitz, Avishai and Shalem, Daniel and Lifschitz, Daphna and Moshe, Dudu and Porat, Eitan and others},
  journal={arXiv preprint arXiv:2601.03233},
  year={2026}
}

@article{wang2026mechanistic,
  title={A Mechanistic View on Video Generation as World Models: State and Dynamics},
  author={Wang, Luozhou and Chen, Zhifei and Du, Yihua and Yan, Dongyu and Ge, Wenhang and Shen, Guibao and Xu, Xinli and Wu, Leyi and Chen, Man and Xu, Tianshuo and others},
  journal={arXiv preprint arXiv:2601.17067},
  year={2026}
}

@article{yue2025simulating,
  title={Simulating the Visual World with Artificial Intelligence: A Roadmap},
  author={Yue, Jingtong and Huang, Ziqi and Chen, Zhaoxi and Wang, Xintao and Wan, Pengfei and Liu, Ziwei},
  journal={arXiv preprint arXiv:2511.08585},
  year={2025}
}

@article{garrido2026learning,
  title={Learning Latent Action World Models In The Wild},
  author={Garrido, Quentin and Nagarajan, Tushar and Terver, Basile and Ballas, Nicolas and LeCun, Yann and Rabbat, Michael},
  journal={arXiv preprint arXiv:2601.05230},
  year={2026}
}

@article{lillemarkflow,
  title={Flow Equivariant World Modeling for Partially Observed Dynamic Environments},
  author={Lillemark, Hansen and Huang, Benhao and Zhan, Fangneng and Du, Yilun and Keller, T Anderson},
  year={2025}
}

@misc{3d_as_code,
    title={3D as code},
    author={World Labs},
    howpublished = {\url{https://www.worldlabs.ai/blog/3d-as-code}},
    year={2025}
}

@article{he2025plaicraft,
  title={Plaicraft: Large-scale time-aligned vision-speech-action dataset for embodied ai},
  author={He, Yingchen and Weilbach, Christian D and Wojciechowska, Martyna E and Zhang, Yuxuan and Wood, Frank},
  journal={arXiv preprint arXiv:2505.12707},
  year={2025}
}

@inproceedings{yu2025gamefactory,
  title={Gamefactory: Creating new games with generative interactive videos},
  author={Yu, Jiwen and Qin, Yiran and Wang, Xintao and Wan, Pengfei and Zhang, Di and Liu, Xihui},
  booktitle={Proceedings of the IEEE/CVF International Conference on Computer Vision},
  pages={11590--11599},
  year={2025}
}

@article{che2024gamegen,
  title={Gamegen-x: Interactive open-world game video generation},
  author={Che, Haoxuan and He, Xuanhua and Liu, Quande and Jin, Cheng and Chen, Hao},
  journal={arXiv preprint arXiv:2411.00769},
  year={2024}
}

@article{nan2024openvid,
  title={Openvid-1m: A large-scale high-quality dataset for text-to-video generation},
  author={Nan, Kepan and Xie, Rui and Zhou, Penghao and Fan, Tiehan and Yang, Zhenheng and Chen, Zhijie and Li, Xiang and Yang, Jian and Tai, Ying},
  journal={arXiv preprint arXiv:2407.02371},
  year={2024}
}

@article{lin2024open,
  title={Open-sora plan: Open-source large video generation model},
  author={Lin, Bin and Ge, Yunyang and Cheng, Xinhua and Li, Zongjian and Zhu, Bin and Wang, Shaodong and He, Xianyi and Ye, Yang and Yuan, Shenghai and Chen, Liuhan and others},
  journal={arXiv preprint arXiv:2412.00131},
  year={2024}
}

@article{ju2024miradata,
  title={Miradata: A large-scale video dataset with long durations and structured captions},
  author={Ju, Xuan and Gao, Yiming and Zhang, Zhaoyang and Yuan, Ziyang and Wang, Xintao and Zeng, Ailing and Xiong, Yu and Xu, Qiang and Shan, Ying},
  journal={Advances in Neural Information Processing Systems},
  volume={37},
  pages={48955--48970},
  year={2024}
}

@article{zhou2025omniworld,
  title={Omniworld: A multi-domain and multi-modal dataset for 4d world modeling},
  author={Zhou, Yang and Wang, Yifan and Zhou, Jianjun and Chang, Wenzheng and Guo, Haoyu and Li, Zizun and Ma, Kaijing and Li, Xinyue and Wang, Yating and Zhu, Haoyi and others},
  journal={arXiv preprint arXiv:2509.12201},
  year={2025}
}

@article{huang2025self,
  title={Self forcing: Bridging the train-test gap in autoregressive video diffusion},
  author={Huang, Xun and Li, Zhengqi and He, Guande and Zhou, Mingyuan and Shechtman, Eli},
  journal={arXiv preprint arXiv:2506.08009},
  year={2025}
}

\end{document}